\documentclass[11pt]{article}
\usepackage{styles/nodalida2025}
\usepackage{float} 
\usepackage{placeins} 
\usepackage{times}
\usepackage{amsmath} 
\usepackage{booktabs}
\usepackage{url}
\usepackage{hyperref}
\usepackage[noabbrev,capitalize,nameinlink]{cleveref}
\usepackage{latexsym}
\usepackage{graphicx}
\usepackage{tablefootnote}
\usepackage{tabularx}
\usepackage[table]{xcolor}
\usepackage{tikz}
\usepackage{tikz-dependency}
\usepackage{xcolor} 
\usepackage{multirow}
\usepackage{caption} 
\usepackage{enumitem}

\newcommand*\circled[1]{\tikz[baseline=(char.base)]{
            \node[shape=circle,draw,inner sep=.6pt] (char) {#1};}}

\title{From Smør-re-brød to Subwords: 

Training LLMs on Danish, One Morpheme at a Time}

\author{
   \textbf{Mikkel Wildner Kildeberg} \hspace{1em} \textbf{Emil Allerslev Schledermann} \hspace{1em} 
    \textbf{Nicolaj Larsen} \\ \textbf{Rob van der Goot}
    \\
    IT University of Copenhagen \\ \texttt{[easc,nicla,mwki]@itu.dk}
}

\date{}
\begin{document}
\maketitle
\begin{abstract}
\looseness=-1
The best performing transformer-based language models use subword tokenization techniques, such as Byte-Pair-Encoding (BPE). However, these approaches often overlook linguistic principles, such as morphological segmentation, which we believe is fundamental for understanding language-specific word structure. In this study, we leverage an annotated Danish morphological dataset to train a semisupervised model for morphological segmentation, enabling the development of tokenizers optimized for Danish morphology. We evaluate four distinct tokenizers, including two custom morphological tokenizers,  by analyzing their performance in morphologically segmenting Danish words. Additionally, we train two generative transformer models, \textit{CerebrasGPT-111M} and \textit{LLaMA-3.2 1B}, using these tokenizers and evaluate their downstream performance. Our findings reveal that our custom-developed tokenizers substantially enhance morphological segmentation, achieving an F1 score of 58.84, compared to 39.28 achieved by a Danish BPE tokenizer. In downstream tasks, models trained with our morphological tokenizers outperform those using BPE tokenizers across different evaluation metrics. These results highlight that incorporating Danish morphological segmentation strategies into tokenizers leads to improved performance in generative transformer models on Danish language\footnote{Repository containing experiments can be found at: \url{https://github.com/emilschleder/DaMorph}}. It should be noted that there has been a concurrent paper with different experiments on the same data~\cite{morsed}.

\end{abstract}

\section{Introduction}

Within Natural Language Processing (NLP) and Large Language Models (LLMs), tokenization serves as a foundational step, determining how textual data is segmented into meaningful units for downstream processing by language models. The performance of LLMs relies on the quality of their tokenization strategies, which influence their ability to understand and generate coherent and contextually relevant text. The most straightforward method of tokenizing a sentence is to split it into words, characters, or segments based on whitespace. However, this approach is inefficient, resulting in either a large vocabulary or insufficient coverage, while also preventing the model from learning from repetitive segments of words.

In order to maintain a large vocabulary and the ability to recognize subword structures, most modern tokenizers rely on \textit{subword tokenization}, which splits words into subwords using various statistical methods \cite{wolleb2024assessingimportancefrequencyversus}. An example could be splitting the word \textit{"unbelievable"} into \textit{"un"}, \textit{"believ"}, and \textit{"able"}. 

Common subword tokenizers, such as Byte Pair Encoding (BPE), WordPiece, Unigram and SentencePiece have become the standard of subword tokenization, demonstrating success across a multitude of languages. Despite the success, their performance often drops when confronted with the unique linguistic and morphological characteristics of specific, lower resource languages (\cref{tab:tokenization_comparison}) \cite{shibata1999byte, sennrich-etal-2016-neural, kudo-2018-subword}. Examples of such morphologically complex languages include Turkish and Swahili, which exhibit extensive word-level complexity. While not as morphologically complex as Turkish and Swahili, Danish still presents challenges for standard tokenization methods due to its compounding and morphological structure. \cite{Toraman_2023, atuhurra2024introducingsyllabletokenizationlowresource, morfemer}.

\begin{table}[t]
\resizebox{\columnwidth} {!}{
\small
\begin{tabular}{l|lll}
\hline
\textbf{Word} & \textbf{Character } & \textbf{Morphemes} & \textbf{Subwords (BPE) } \\ \hline
\textit{kranie} & k r a n i e & kranie & k-ran-ie \\ \hline
\textit{landstræner} & l a n d s t r æ n e r & land-s-træn-er & land-str-æ-ner\\ \hline
\textit{lånte} &   l å n t e  & l-å-nte & l-å-nte\\ \hline
\textit{skoletaske} &    s k o l e t a s k e  & skole-taske & sko-let-as-ke\\ \hline
\textit{venlig} &   v e n l i g & ven-lig & ven-lig\\ \hline
\end{tabular} }
\caption{Examples of tokenization methods.}
\label{tab:tokenization_comparison}
\end{table}

In the field of linguistics, morphemes are described as the smallest meaningful units of text in natural language \cite{sinclair1996search, linguisticsMorphologyBook}. In that sense, morphemes can be characterized as meaningful subwords, which present a promising alternative to the established subword tokenization approaches, as morphemes are based on linguistic characteristics rather than statistical co-occurrences of characters.

Danish demonstrates morphological complexity, particularly through its frequent use of compound words formed by concatenating multiple morphemes \cite{morfemer}. For instance, the Danish word \textit{``Smørrebrød''} combines \textit{``Smør''} (eng: butter) and \textit{``brød''} (eng: bread) using the \textit{``-re-''} as link to denote a traditional open-faced sandwich. Inadequate handling of such compound formations can lead to morphologically suboptimal tokenization, as common tokenizers may either split compounds inappropriately or fail to recognize them as morphologically meaningful segments. This may lead to fragmented tokens that can hinder the model's ability to generate accurate representations of the language \cite{hofmann-etal-2022-embarrassingly, hofmann-etal-2021-superbizarre}. 

These concerns highlight an unresolved potential within Danish NLP, where the lack of morphologically optimized tokenization strategies for Danish hinder the development of language models for this language. An important part in why this potential still remains unresolved is that resources containing Danish morphemes are very limited, making it difficult to train even a small tokenizer optimized for Danish morphological segmentation. 

This gap in resources underscores the need for innovative approaches to tackle the challenges of Danish morphological segmentation. To address this, we make the following contributions: \circled{1} We evaluate semi-supervised and unsupervised Morfessor models trained on both annotated and non-annotated data. \circled{2} We develop two custom Hugging Face PreTrainedTokenizer subclasses tailored for Danish morphological segmentation. \circled{3} We conduct experiments with our developed tokenizers to evaluate their performance on morphological segmentation, ensuring their effectiveness for Danish linguistic tasks. \circled{4} We assess the performance of our custom tokenizers in training generative transformer models, using \textit{LLaMA 3.2 1B} and \textit{Cerebras-GPT 111M}. \circled{5} We initiate a full fine-tune of \textit{LLaMa 3.2 3B} with the best performing tokenizer on a large Danish text corpus.

\section{Related Work}
Statistical approaches to morphological segmentation encompass both unsupervised techniques, such as those outlined in \cite{virpioja-etal-2011-empirical} which detect statistical patterns in language data, and methods like Morfessor, which leverage probabilistic modeling and facilitate both unsupervised and semi-supervised learning \cite{virpioja2013morfessor, goldsmith}. More recently, neural-based approaches have become popular morphological segmentation techniques. Examples include Long short-term memory (LSTM) architectures, which are designed to learn word structures and predict morphological boundaries, and neural attention models that leverage attention mechanisms to effectively identify morphological segmentation in a supervised fashion \cite{Wang_Cao_Xia_de_Melo_2016, Zhu2018}. 

Despite these approaches, morphological segmentation in the context of language modeling is still a relatively unexplored area, especially for lower-resource languages such as Danish. Morphological segmentation has previously been applied in machine translation with mixed results \cite{salameh-etal-2013-reversing, pan-etal-2020-morph-egment, DBLP:journals/corr/abs-1806-05482, banerjee-bhattacharyya-2018-meaningless}. Still, early contributions suggest that morphologically informed subword tokenizers may improve language modeling performance \cite{bostrom-durrett-2020-byte}.

Implementing morphological segmentation in subword tokenizers remains relatively limited in scope, with most studies focusing on languages like English and Turkish. Some recent advancements include the Unsupervised Morphological Tree Tokenizer \cite{zhu2024unsupervisedmorphologicaltreetokenizer}, a Turkish morphological-level tokenizer \cite{Toraman_2023}, and MorphPiece — a linguistically informed tokenization method combining morphological segmentation and BPE, presenting concurrent work with our research \cite{jabbar2024morphpiecelinguistictokenizer}. However, MorphPiece differentiate from our research by not utilizing Morfessor and by relying solely on annotated data from the MorphNet dataset \cite{batsuren-etal-2021-morphynet}. More recently, MYT5 was introduced shortly after our experiments, incorporating Morphology-Driven Byte Encoding (MYTE), an encoding approach based on morphemes generated using the Morfessor framework. By leveraging morphemes, MYT5 produces shorter and more meaningful encodings \cite{Limisiewicz_2024}. These works highlight the emerging interest in linguistically motivated tokenization schemes for improved model performance.

Our research builds on this foundation by developing a Danish-specific subword tokenizer, leveraging the Morfessor framework \cite{virpioja2013morfessor}. While prior studies have explored language-specific tokenization, our work integrates Morfessor specifically for Danish morphological structure. Our approach is similar to MorphPiece in its combination of utilizing a morph-table in combination with BPE; however, it diverges in several key aspects. First, our focus is exclusively on Danish, and second, we employ Morfessor to address Danish morphology on downstream evaluation tasks. To the best of our knowledge, our implementation is the first to apply the Morfessor framework for morphological segmentation within a Danish language modeling context.

\section{Research Question}

This study is guided by the following research question, which aims to explore the impact of morpheme tokenization techniques on language model performance in Danish:
\begin{quote}
\textit{To what extent can morpheme-based subword tokenization enhance the linguistic and grammatical coherence of Danish text produced by transformer-based language models compared to a widely used subword tokenization method?}

\end{quote}

To address this research question, we have formulated a set of sub-questions to guide and support our investigation:

\begin{enumerate}
  \setlength\itemsep{0cm}
    
    \item What effects do semi-supervised and unsupervised learning have on the Morfessor segmentation accuracy in Danish?
     
    \item How does the performance of a morphologically informed tokenizer compare to common subword tokenizers in accurately segmenting Danish words into morphemes?
     
    \item How does a morphologically informed tokenizer impact generative transformer model performance in downstream tasks for Danish?
\end{enumerate}

\section{Setup \& Methods}

\begin{table}[t]
    \centering
    \small
    \resizebox{\linewidth}{!}{
    \begin{tabular}{l|lll }
    \toprule
    \textbf{Dataset}   & \textbf{Domain} & \textbf{Words} & \textbf{Source}  \\
    \midrule
    Bookshop  & Books    & 208M  & \citet{tiedemann-2012-parallel}                        \\
    CC-100     & Webscrape & 7.82B & \citet{wenzek-etal-2020-ccnet}                            \\
    CulturaX    & Webscrape & 14.8B  & \citet{nguyen2023culturax}                               \\
    DaNewsroom  & News & 391M  & \citet{varab-schluter-2020-danewsroom}                   \\
    FTSpeech    & Political & 43.3M  & \citet{kirkedal2020ft}                                  \\
    Gigaword     & Mixed & 1.02B  & \citet{stromberg-derczynski-etal-2021-Danish}          \\
    OpenSubtitles & Subtitles & 207M &  \citet{lison-tiedemann-2016-opensubtitles2016}    \\
    Reddit        & Social & 73.9M & \citet{chang-etal-2020-convokit}                           \\
    Twitter       & Social & 21.9M & \scriptsize{\url{archive.org/details/twitterstream}}   \\
    Dawiki      & Wiki & 62.4M & \citet{Wikiextractor2015}                                 \\
    \midrule
    \textbf{Total} & & \textbf{24.6B} & \\
    \bottomrule 
    \end{tabular}
    }
    \caption{List of datasets.}
    \label{tab:data}
\end{table}

\subsection{Datasets} 
For training the Morfessor algorithm and BPE tokenizer we use text data from a variety of sources. The BPE tokenizer is trained on a combination of all datasets (24,6 billion words) (\cref{tab:data}). For Morfessor we use a dataset of 821 words and their correctly segmented morphemes, annotated by a domain expert. To further optimize the morpheme segmentation, we experiment with multiple datasets, including the 400.000, 2 million and 10 million most frequent words across all sources (\cref{tab:data}) and a dataset of 400.000 unique words from Dansk Sprognævn (COR) \footnote{\url{https://dsn.dk/}}. Due to computation and time limitations, our generative transformer models are trained on 25\% of the Gigawords dataset - a comprehensive Danish corpus consisting of one billion Danish words \cite{stromberg-derczynski-etal-2021-Danish}. We shuffled the data randomly to ensure representation from all sources.

\begin{table*}[t]
\centering
\resizebox{\linewidth}{!}{
    \begin{tabular}{l|lll}
    \toprule
    \textbf{Word} & \textbf{Translated (eng.)} & \textbf{Category} & \textbf{Split}  \\
    \midrule
    Kranie & Skull  & Root Morpheme & Kranie	[Root] \\ 
    Landstræner & National coach & Linking Morpheme & Land[Root] s[Link] træn[Root] er[Suff] \\
    Lånte & Borrowed & Inflection & Lån[Root] te[Infl] \\
    Bibringe & Impart & Prefix & Bi[Pref] bringe[Root] \\
    Skoletaske & School bag & Compound & Skole [Root] taske [Root]  \\ 
    Venlig & Friendly & Suffix & Ven[Root] lig[Suff] \\
    \midrule
    \end{tabular}
    }
    \caption{Example words of each segmentation category.}
    \label{tab:morphemes}
\end{table*}

\subsubsection{Annotated morphemes} 
As previously mentioned, no public Danish morpheme dataset exists. Therefore, we collaborated with a Danish expert annotator, who created a small morpheme dataset comprising words into categories of root morphemes, compounds, compounds with linking elements, prefixes, suffixes, and inflections. In \cref{tab:morphemes} one word of each category is represented. The dataset consist of 821 total words, 269 root morphemes, 225 compounds, 98 compounds with linking, 82 prefixes, 82 suffixes and 64 inflections.

The annotator has 35 years of experience as a Danish teacher, holds a postgraduate diploma in “Reading for Adults” and has extensive expertise in teaching individuals with difficulties in written language. Her work focuses on the use of morphemes as a foundational tool for improving language comprehension. Furthermore, she has authored a book on this exact topic \cite{morfemer}, underscoring her authority in relation to this project.

\textit{Root morphemes} are fundamental units, carrying the core meaning part of a word and is capable of standing alone as an independent word. \textit{Compounds} are formed by combining two or more root morphemes, with the final morpheme often denoting the primary category, and the preceding elements denoting its type. \textit{Linking morphemes}, such as "e" and "s" serve as connecting morphemes between root morphemes in compounds. \textit{Prefixes} are the starting morphemes, attached to the beginning of a root, to alter its meaning without changing the grammatical category. \textit{Suffixes} are ending morphemes, attached to the end of a root morpheme and often modifies its meaning and grammatical category. Lastly, \textit{Inflections} are bound morphemes that indicate grammatical relationship, such as gender, number and tense, without altering the root morphemes' meaning. They are mainly associated with verbs, nouns and adjectives. Together, the mentioned categories are important aspects of the Danish language construction and syntax \cite{morfemer}. 

\subsection{Segmentation methods}

\subsubsection{Morfessor} 

The Morfessor framework is a morphological segmentation tool, facilitating morpheme-based subword tokenization for NLP tasks.

Utilizing probabilistic models, Morfessor identifies and segments words based on the statistical properties of character sequences within a given corpus, without the necessity for annotated data. This capability makes Morfessor particularly advantageous for languages with small-to-none annotated data possibilities and morphological structures \cite{smit-etal-2014-morfessor}.  

Morfessor supports both supervised, unsupervised and semi-supervised learning approaches. For this project, we have utilized the semi-supervised and unsupervised approaches. The semi-supervised approach allows us to provide the model with a set of annotated data to affect the model calculations in combination with a non-annotated large text corpus. In contrast, the unsupervised approach allows the model to optimize its calculations based solely on the non-annotated corpus \cite{smit-etal-2014-morfessor}.

In order to determine the optimal Morfessor model to integrate into our tokenizers, we performed a series of experiments as presented in \textit{\autoref{sec:best_morf}}.

\begin{table*}[t]
\centering
\resizebox{\linewidth}{!}{
    \begin{tabular}{l|llllll}
    \toprule
    \textbf{Model}  & \textbf{Parameters} & \textbf{Tokenizer type} & \textbf{Vocab size} & \textbf{Sequence length (tokens)} &  \textbf{Multilingual} \\
    \midrule
    LLaMa 3.2 1B & 1.231.000.000 & BPE & 128.256 & 16.000 &  Yes \\ 
    Cerebras-GPT 111M & 111.000.000 & BPE & 50.257 & 2.048 &   No \\
    \midrule
    \end{tabular}
    }
    \caption{Models overview.}
    \label{tab:selected_models}
\end{table*}

\subsubsection{BPE} 
We adopt BPE as introduced by \cite{shibata1999byte}, which merges the most frequent pairs of characters into subword units iteratively until a predetermined vocabulary size is achieved. BPE is a common subword tokenization technique that balances the granularity of tokens, enabling the handling of out-of-vocabulary words by decomposing them into more manageable subword components. The algorithm operates by initially treating each character as an individual token and progressively merging the most frequent adjacent token pairs. This process continues until the fixed vocabulary size is reached, resulting in a set of subword tokens that capture common patterns within the corpus \cite{shibata1999byte}.

\begin{figure*}[t]
    \centering
    \includegraphics[width=1\textwidth]{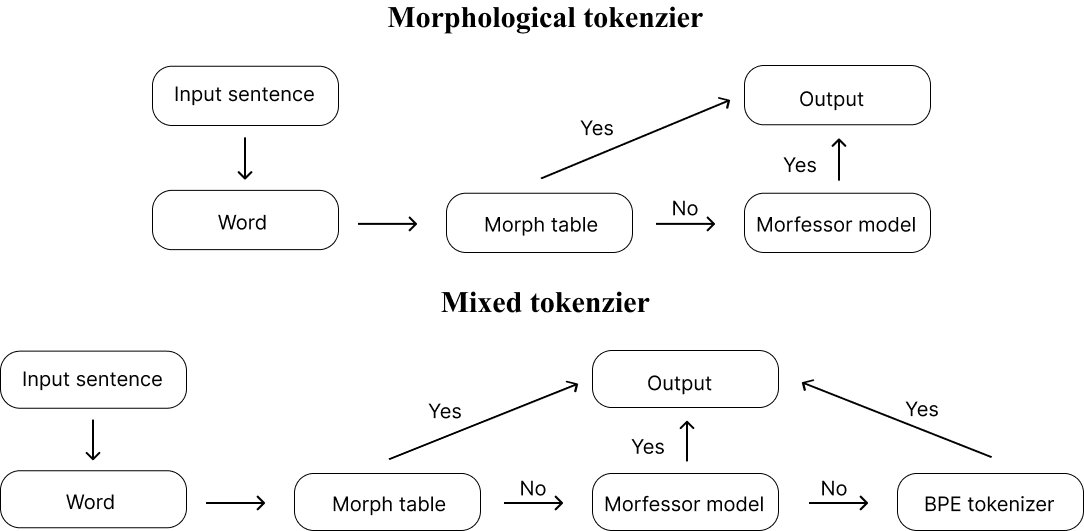}
    \caption{Morphological PreTrainedTokenizer implementation. The tokenizer prioritizes morpheme segmentation, then defaulting to the Morfessor model. The Mixed tokenizer uses BPE as a fallback if the Morfessor split is not found in the vocabulary}
    \label{fig:Token}
\end{figure*}

\subsection{Custom Morphological tokenizer}
To leverage Morfessor in a language model, we developed a tokenizer based on the PreTrainedTokenizer class from the Hugging Face tokenizers library. To effectively train the  tokenizer, we constructed a morpheme table (morphtable) that included words and their corresponding morpheme segments in the vocabulary. This table was created by identifying the most frequent "chunks" in the combined dataset and applying Morfessor until we reached the desired vocabulary size. By “chunk,” we refer to the combination of a leading whitespace, a word, and any special characters. This allows Morfessor to handle the splitting of whitespace and special characters. An example of this is the sentence \textit{“Hej med dig!”} (eng: \textit{“Hello there!”}), which would be split into \textit{“Hej”}, \textit{“ med”}, and \textit{“ dig!”}. This approach was implemented to enhance the efficiency of tokenization, as relying solely on the Morfessor model resulted in longer processing times. The tokenizer continued to rely on the Morfessor model, but only when it could not locate the word in the morpheme table. The tokenizer integrates two implementations:

\begin{itemize}
    \item A \textbf{Morphological tokenizer} that relies sorely on a morphtable and Morfessor model.
    \item A \textbf{Mixed tokenizer} utilizing both a morphtable, a Morfessor model and BPE.
\end{itemize}

The Mixed tokenizer contains a vocabulary file with morphological segments (a morph table), a Morfessor model, a BPE vocab and special tokens, allowing for efficient word segmentation. During encoding, the tokenizer first attempts to segment words into morphemes using the morphtable. If no segments are found, it attempts to segment with the Morfessor model. If segmented morphemes are not available in the vocabulary, it defaults to BPE tokenization as presented in \cref{fig:Token}. 

\subsection{Model selection}

We chose to train two models: \textit{Cerebras-GPT 111M} and \textit{LLaMA 3.2 1B} (\cref{tab:selected_models}). Both models are generative transformer models, but not instruction-tuned, which means that they are unable to follow instructions provided by the user. Instead, the two models are text-generation models, meaning that they are capable of generating new text based on a text-input - e.g. to continue a sentence.

The Cerebras-GPT 111M model was selected for its manageable size and lower computational demands. It is a generative transformer model with 111 million parameters based on the GPT3 architecture. The model has been trained with a maximum sequence length of 2.048 tokens and has a vocabulary size of 50.257. It has been pretrained on the pile dataset which consist of 22 data sources and is primarily focused on English \cite{dey2023cerebrasgptopencomputeoptimallanguage}. 

In comparison, the LLaMa 3.2 1B is much larger with a total of 1 billion parameters - approximately nine times the size of the Cerebras model. The LLaMa model has a vocabulary size of 128.256 and a maximum sequence length of 16.000 tokens. It has been pretrained on a massive dataset of 15 trillion tokens, which may include some exposure to Danish \cite{dubey2024llama3herdmodels}. LLaMA 3.2 1B was chosen for its state-of-the-art performance, but due to time constraints, we limited the training of Llama 3.2 1B to the embeddings layer. This also allowed us to investigate whether focusing exclusively on embedding training with a Danish subword tokenizer could still improve performance. However, we hypothesize that only training the embeddings will have an impact on model performance compared to conducting a full model training.

\subsection{Training}

\begin{table}[t]
    \centering
    \small
    \resizebox{\linewidth}{!}{
    \begin{tabular}{l|ll}
    \toprule
        \textbf{Model} & \textbf{Tokenizer Type} & \textbf{Vocab size} \\
    \midrule
        \multirow{4}{*}{LLaMA 3.2 1B} 
        & Standard BPE Tokenizer & 128.256 \\
        & Danish BPE Tokenizer & 128.256 \\
        & Morphological Tokenizer & 128.256 \\
        & Mixed Tokenizer & M: 76.956 B: 51.300 \\
        \hline
        \multirow{4}{*}{Cerebras-GPT 111M}
        & Standard BPE Tokenizer & 50.257 \\
        & Danish BPE Tokenizer & 50.257 \\
        & Morphological Tokenizer & 50.257 \\
        & Mixed Tokenizer & M: 30.157 B: 20.100 \\
    \bottomrule
    \end{tabular}
    }
    \caption{Overview of vocab sizes. M = morphological vocab, B = BPE vocab.}
    \label{tab:model_selection}
\end{table}

Both models were trained on either 4x A100 GPUs or 6x v100 GPUs for up to seven days. We employed mixed-precision training for both Cerebras-GPT 111M and LLaMA 3.2 1B, utilizing the Adam optimizer. We opted to use the same learning rate as specified in the official training paper for each model, setting it to 6 $\times$ 10\textsuperscript{$-4$} for Cerebras-GPT 111M and 3 $\times$ 10\textsuperscript{$-4$} for Llama 3.2 1B \cite{dey2023cerebrasgptopencomputeoptimallanguage, dubey2024llama3herdmodels}. We configured the training process with a batch size of 1 for LLaMA 3.2 1B and 18 for Cerebras-GPT 111M. Training was conducted for a single epoch on 25\% of the Gigawords dataset. We conducted eight separate training sessions across two models: four LLaMA 3.2 1B (only embeddings layer) and four Cerebras-GPT 111M. We initialize the model and embedding weights based on the Xavier uniform distribution \cite{xavier-dist}.
For each model, the training configurations consisted of: 

\begin{itemize}
    \item \textbf{Standard Tokenizer}: Trained with the default tokenizer.
    \item \textbf{Danish BPE Tokenizer}: Trained with a Danish-specific BPE tokenizer.
    \item \textbf{Morphological Tokenizer}: Trained with our morphological tokenizer, leveraging a Morfessor model and a customized morpheme table.
    \item \textbf{Mixed Tokenizer}: Trained with a combination of the morphtable, the Morfessor model and a Danish-Specific BPE tokenizer.
\end{itemize}

We chose to retain the models default vocabulary sizes for all experiments to keep the number of experiments minimized. While it could be argued that these vocabulary sizes are large for a single language, keeping them allows for a fair comparison with models trained using their standard tokenizers. Moreover, we observed that our tokenizers with the largest vocabulary sizes yielded the highest F1 scores for morpheme-based word segmentation, as detailed in the results section. For the mixed tokenizers we reserved 60\% of the vocab size for the morphtable and 40\% for BPE as shown in \cref{tab:model_selection}. This distribution was chosen to ensure that the tokenizers primarily relied on the morphtable. This approach aligns with MorphPiece, but with a different ratio between the BPE and Morphologically informed vocabularies \cite{jabbar2024morphpiecelinguistictokenizer}.

\begin{table*}
\centering
\resizebox{\linewidth}{!}{
\small
    \begin{tabular}{l|llll}
    \toprule
    \textbf{Category}  & \textbf{Story} & \textbf{TopP} & \textbf{TopK} & \textbf{Temp}  \\
    \midrule
    Conditional Sentence & "Hvis jeg havde vidst, at det ville regne hele dagen, ville jeg have..." & 0.85 & 30 & 0.7 \\ 
    Dialog with Emotion & "Hvorfor i alverden gjorde du det? Jeg kan ikke forstå, at du..." & 0.85 &30 & 0.8\\
    Descriptive Language  & "Den gamle kro ved havnen var kendt for sin hyggelige atmosfære og…" & 0.9 & 40 & 0.85\\ 
    Historical Narratives  & "For mange år siden, da jeg var en ung dreng, plejede jeg at…" & 0.95 & 50 & 0.9\\
    \midrule
    \end{tabular}
    }
    \caption{The 4 prompts for text generation and the parameters for the model.}
    \label{tab:prompts}
\end{table*}
\subsection{Metrics for evaluation.}
For evaluating our results, we use four complementary evaluation methods: 

\begin{itemize}
    \item An intrinsic evaluation using F1 scores to asses the performance of our tokenizers.
    \item An intrinsic measure, consisting of Bits Per Character (BPC) and Bits Per Token (BPT), to assess model efficiency.
    \item An extrinsic benchmark, ScandEval, to evaluate the model's downstream performance.
    \item A qualitative analysis with five Danish native speakers to provide human-centered insights into the language model's capabilities.
\end{itemize}

\subsubsection{BPC and BPT}
BPC and BPT are metrics used in language modeling to calculate the average number of bits required to encode each character (BPC) and token (BPT) in a text. The metrics are derived from the cross-entropy loss and is calculated by:

\[
\text{BPC} = \frac{-\sum_{i=1}^N \log_2 P(x_i)}{N_{characters}}
\]

\[
\text{BPT} = \frac{-\sum_{i=1}^N \log_2 P(x_i)}{N_{tokens}}
\]

Where $N_{tokens}$ and $N_{characters}$ are the total number of tokens or characters in the dataset and $P(x_i)$ is the probability. A lower score could indicate more efficient and accurate language models, as they reflect a higher degree of predictability and compression of character/token sequences. If a model achieves a score of 1, it means that on average, 1 bit is needed to encode each character/token. A higher score indicates lower compression efficiency, as more bits are required. To calculate the score, we evaluate the model's ability to predict the next character/token in a sequence. We calculated BPC and BPT on the Twitter dataset, as shown in \cref{tab:data}, which neither the tokenizer nor the models have been exposed to.                                                                                                                

\subsubsection{ScandEval}
ScandEval is a comprehensive evaluation framework designed to assess the quality of generated text through a series of semantic analyses for Nordic languages. This framework encompasses various sub-evaluations, including grammatical correctness, coherence, and contextual relevance, thereby offering an evaluation of the language model’s performance. We focused our ScandEval evaluation on the \textit{Linguistic Acceptability} (LA) and \textit{Summarization} (SUM) sub-evaluations, as the other sub-evaluations are primarily designed for large language models with advanced language proficiency that have been trained on much larger datasets. 

\textbf{LA} is a Danish binary classification evaluation where a model is tasked to predict whether or not a sentence is grammatically correct. The evaluation is measured by two metrics, the F1 score and Matthew’s Correlation Coefficient (MCC). A higher F1 score reflects better prediction accuracy. While the MCC ranges from -100\% to 100\%, the closer the score is to 100\%, the more reliable and deterministic the model’s predictions are, as opposed to being random.  

\textbf{SUM} is a task where the model is provided with a text and tasked with generating a summary. The texts to be summarized are primarily Danish news articles. The performance is then evaluated using BERT and ROUGE-L scores. BERT is estimating the performance by looking at semantic similarities whereas ROUGE-L focuses on the longest matching sequences of words between the models output and the reference text \cite{nielsen2023scandeval}. Although we anticipate that the SUM metric may not be the most critical in evaluating the models, we include it because it is the only metric that assesses the generative model’s output. In contrast, the LA metric is based on predicting a single label.

\subsubsection{Human Qualitative Evaluation}

In addition to our quantitative evaluations, we conducted a qualitative study to further assess the models' performance. Five native Danish speakers participated in this evaluation, where they were tasked with evaluating text generation outputs produced by the models under our four distinct tokenization conditions. We employed four varied prompts, each focusing on a specific theme: a conditional sentence, an emotional dialogue, descriptive language, and a historical narrative. We decided to focus mainly on the Cerebras models, as these underwent full parameter fine-tuning. We also included a single LLaMa model, the \textit{Da. Morph. LLaMa}, as the tokenizer used in this model showed the best performance in Morphological segmentation (\cref{tab:combined_metrics}). For each prompt, the models generated continuations, resulting in a total of 20 generated texts. The prompts can be found in \cref{tab:prompts}. 

Participants were provided with the prompts and the corresponding continuations. They were instructed to assess each continuation based on three primary criteria:

\begin{enumerate}
    \item \textbf{Fluency}: Measuring the smoothness and readability of the generated text in isolation.
    \item \textbf{Grammar}: Evaluating the correctness of language use.
    \item \textbf{Naturalness}: Assessing how naturally the continuation matches the prompt.
\end{enumerate}

Ratings were collected using a Likert scale consisting of: 1 (Incomprehensible), 2 (Barely Understandable), 3 (Somewhat Understandable), 4 (Mostly Understandable) and 5 (Distinguishable) for each criterion. We randomized and anonymized the models to ensure they were evaluated without prior knowledge of their identity.

As seen in \cref{tab:prompts} we used different settings for top p, top k and temperature. Using higher values for top-k, top-p, and temperature encourages more creative and descriptive outputs, while lower values tend to produce more precise and focused responses. For “Dialog with Emotion” and “Conditional Sentence,” we opted for slightly lower values to enhance relevance and minimize overly random outputs. In contrast, for the remaining two categories, we selected higher values to enable the models to generate richer and more engaging language. We also chose to limit the repeating n-gram size to a maximum of 3 to avoid repetitive phrases or words.

\begin{figure*}[t]
    \centering
    \includegraphics[width=1\textwidth]{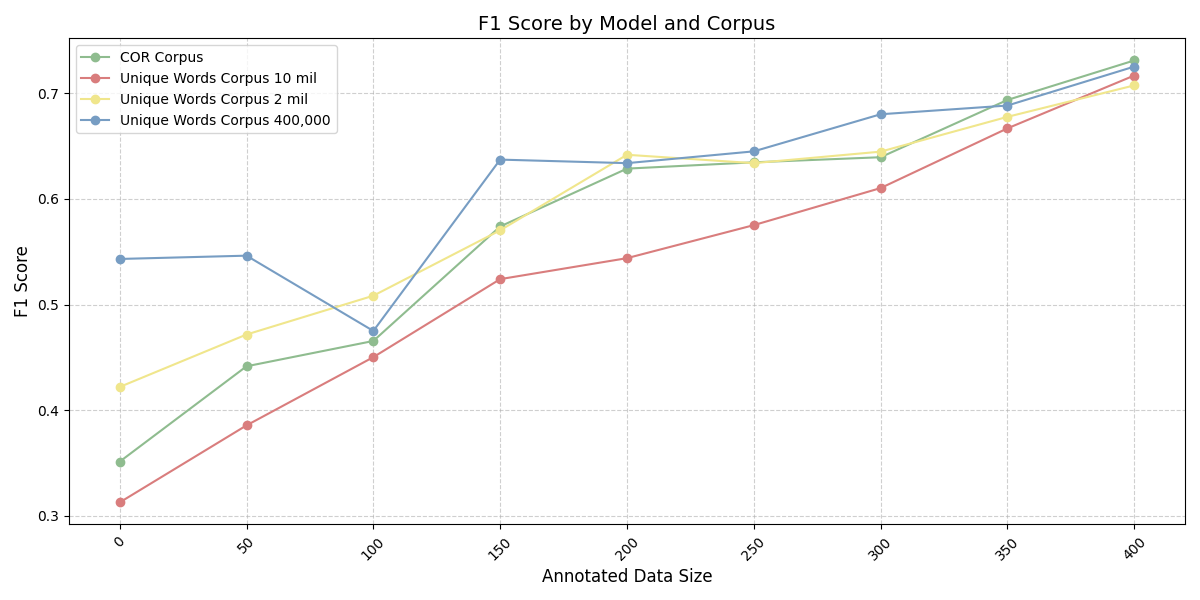}
    \caption{F1 score for each Morfessor model and their amount of annotated data. F1 counts the correct segments.}
    \label{fig:data_experiments}
\end{figure*}

\section{Results}

\subsection{Finding the best Morfessor model}
\label{sec:best_morf}
In order to find the most optimal Morfessor model to integrate into our tokenizer, we began by conducting experiments relying on unsupervised learning by using four different data samples: 400.000 words from the COR dataset along with the 400.000, 2 million and 10 million most frequent words across all datasets in \cref{tab:data} (excluding twitter). Following the unsupervised approach, we then utilized the Morfessor framework to extend the models using annotated data, enhancing their segmentation accuracy through semi-supervised learning. Specifically, we experimented with eight different annotation ratios — ranging from 50 to 400 annotations in increments of 50 to evaluate the impact of varying degrees of annotated data. 

We set aside 71 of the annotated morphemes for evaluation purposes and calculated the recall, precision, F1 and accuracy scores. This evaluation was performed on all Morfessor models based on both annotated and non-annotated data (\cref{fig:data_experiments}). This limited evaluation consisting of 71 words was due to the dataset being finalized shortly before the submission deadline, as the annotation process was time consuming. The results shows that the addition of annotated data substantially increase the models' ability to segment words into morphemes. However increasing the size of the non-annotated sample dataset does not improve performance. Instead, the highest performance is achieved with the 400.000 words from the COR dataset and 400 annotated words. This highlights the importance of an annotation dataset and quality over quantity in non-annotated data samples. The Morfessor model with 400 annotations trained on the 400.000 words from the COR dataset achieved an F1 score of 0.73, outperforming the model with no annotated data, that attained an F1 score of 0.35 (\cref{fig:data_experiments}).

\subsection{Results on morphological tokenization}

To evaluate our tokenizers, we conducted experiments assessing their effectiveness in morphological segmentation of Danish, by using 421 words with their correct morphological segmentations from our annotated dataset. The evaluation dataset consist of 145 root morphemes, 102 compounds, 63 compounds with linking, 52 prefixes, 33 suffixes and 26 inflections. The evaluation measured Precision, Recall, and F1 Score for each tokenizer, along with per-category F1 Scores for Root morphemes, Compounds, Compounds with Linking elements, Prefixes, Suffixes, and Inflections. Results are shown in \cref{tab:combined_metrics}.

The results demonstrate variability in tokenizer performance. Notably, the \textit{Da. Morph. LLaMa} tokenizer achieved the highest overall F1 score of 58.84, while the \textit{Da. Morph. Cerebras} tokenizer recorded the top score among the Cerebras tokenizers with an F1 score of 58.05, highlighting their effectiveness in capturing Danish morphological patterns. Tokenizers integrating both Morfessor and BPE, such as \textit{Da. Mixed. LLaMa} and \textit{Da. Mixed. Cerebras}, achieved a slightly lower F1 Score of 56.83 and 52.78, respectively. This suggests that combining Morfessor and BPE can enhance morphological segmentation, though not surpassing the performance of Morfessor-only tokenizers.

\begin{table*}
\centering
\resizebox{\linewidth}{!}{
\small
\begin{tabular}{l|rrr|rrrrrr|rrrr|r}
\toprule
&  \multicolumn{3}{c|}{\textbf{Vocab Size}} &\multicolumn{6}{c|}{\textbf{Categories F1}} &\multicolumn{4}{c|}{\textbf{Average}} & \textbf{Correct (\%)} \\
\textsc{\textbf{Tokenizer}} & Total & Morph & BPE   & Root & Comp. &  Link. & Pref. & Suff. & Infl. & $P$ & $R$ & $F_1$ & Acc. & Matches \\
\midrule
Std. BPE. Cerebras & 50,257 & \cellcolor{gray!25} & 50,257   & 16.51 & 10.78  & 7.55  & 22.48 & 29.59 & 22.09 & 15.20 & 22.94 & 18.17 & 12.02 & 13.21 \\
Da. BPE. Cerebras & 50,257 & \cellcolor{gray!25}& 128,256    & \textbf{86.90} & 49.05 & 19.89  & 23.42 & 11.76  & 8.68  & 34.72 & 32.39 & 33.27 & 24.39 & 41.04 \\
Da. Morph. Cerebras & 50,257 &  50,257 &  \cellcolor{gray!25} & 82.76 & \textbf{56.57} & \textbf{41.08} & \textbf{41.18} & \textbf{69.35} & \textbf{57.38} & \textbf{71.12} & 50.37 & \textbf{58.05} & 44.35 & 51.65 \\
Da. Mixed. Cerebras & 50,257  &  29,257 & 21,000   & 76.19 & 54.03 & 38.74 & 37.39 & 61.43 & 48.89 & 51.33 & \textbf{54.66} & 52.78 & \textbf{45.32} & \textbf{52.59} \\
\hline
Std. BPE. LLaMa & 128,256 & \cellcolor{gray!25}  & 128,256      & 19.37 & 13.29  & 9.52 & 22.76 & 36.81 & 15.69 & 16.81 & 23.80 & 19.57 & 13.75 & 15.33 \\
Da. BPE. LLaMa & 128,256 & \cellcolor{gray!25} &128,256  &  \textbf{85.03} & \textbf{60.85} & 26.74  & 29.73 & 22.76 & 10.00  & 40.78 & 38.37 & 39.28 & 30.30 & 46.70 \\
Da. Morph. LLaMa & 128,256 & 128,256  & \cellcolor{gray!25} & 83.87 & 55.80 & \textbf{41.42} & \textbf{40.66} & \textbf{69.29} & \textbf{59.20} & \textbf{68.28} & 52.93 & \textbf{58.84} & 47.77 & 55.66 \\
Da. Mixed. LLaMa & 128,256 & 76,956 & 51,300      & 84.59 & 58.89 & 41.07 & 37.86 & 65.67 & 51.91 & 58.44 & \textbf{55.70} & 56.83 & \textbf{48.67} & \textbf{57.31} \\
\bottomrule
\end{tabular}
}
\caption{Morphological segmentation evaluation table for tokenizers.}
\label{tab:combined_metrics}
\end{table*}
The results also shows that both the Morph and Mixed tokenizers outperform the BPE tokenizers. The Morph tokenizer is particularly effective at segmenting words into distinct categories. However, an interesting observation is that the Mixed tokenizer achieves a slightly higher number of correct matches. This suggests that while the Morph tokenizer is generally better at segmenting words, it does not always produce completely accurate splits. For instance, when splitting the word \textit{landshold} (Correct Split: "land-s-hold", eng: "national team") into “land-s-ho-ld,” it successfully identifies “land” and “s” but may not perfectly handle all segments. 

In contrast, tokenizers based solely on BPE, including both standard and Danish-specific models, demonstrated lower performance, underscoring the limitations of BPE in effectively capturing Danish morphological structures (Std. BPE and Da. BPE). The results show that the \textit{Da. Morph. Cerebras} and \textit{Da. Morph. LLaMa} tokenizers achieved the highest F1 scores in most categories (5 and 4 out of 6, respectively). An interesting observation is that the \textit{Da. BPE Cerebras} and \textit{Da. BPE LLaMa} tokenizers performs slightly better on root morphemes compared to the morphological tokenizers. The results also reveal that both the \textit{Da. Morph} and \textit{Da. Mixed} tokenizers perform comparatively worse on compounds with linking elements and prefixes than on other categories, a finding that will be examined further in the analysis.

Regarding the precision and recall scores, both the \textit{Std. BPE.} and \textit{Da. BPE.} tokenizers perform poorly, indicating that they miss many correct tokens. While the \textit{Da. BPE.} shows improvement over the standard BPE, the morphological tokenizers generally achieve higher and more balanced scores. Among these, the \textit{Da. Morph.} tokenizers exhibit slightly higher precision, suggesting that they are more conservative. However, this comes at the cost of missing more correct tokens compared to the \textit{Da. Mixed.} tokenizers. It is also important to note that the results in \cref{tab:combined_metrics} demonstrate that the \textit{Da. Mixed.} and \textit{Da. Morph.} tokenizers perform very similarly. This confirms that there are relatively few instances where the tokenizer defaults to BPE. 

\begin{table*}[t]
\centering
    \resizebox{\linewidth}{!}{
    \begin{tabular}{l|cc|cc|cc}
   \toprule
     \multicolumn{1}{c}{} & \multicolumn{2}{|c}{} & \multicolumn{2}{|c|}{\textbf{LA}} & \multicolumn{2}{c}{\textbf{SUM}} \\
    \textbf{Model} & BPC & BPT & F1 & MCC (\%) &  BERT & ROUGE-L   \\
    \midrule
    Std. BPE. Cerebras & 3.13 & \textbf{6.32} & 33.47 ± 0.33 &-0.86 ± 0.77 & 17.07 ± 2.11 & 2.13 ± 0.26   \\ 
    Da. BPE. Cerebras & 3.08 & 8.48 & 43.51 ± 2.85 & -1.74 ± 1.68  & \textbf{33.99 ± 1.24} & \textbf{2.15 ± 0.09}   \\ 
    Da. Morph. Cerebras & \textbf{3.01} & 6.87 & \textbf{49.61 ± 0.50} & -0.21 ± 1.04 & 10.96 ± 1.99 & 1.15 ± 0.26 \\
    Da. Mixed. Cerebras & 3.34 & 7.50 & 45.84 ± 2.39 & 0.51 ± 1.10 & 16.52 ± 3.40 & 1.09 ± 0.25 \\
            \hline
    Std. BPE. LLaMa & \textbf{4.02} & \textbf{9.40} & \textbf{49.29 ± 0.62} & 0.05 ± 1.33 & 29.37 ± 3.81 & \textbf{2.13 ± 0.33} \\ 
    Da. BPE. LLaMa & 4.59 & 13.45 & 33.28 ± 0.27 & 0.23 ± 0.45 & \textbf{36.54 ± 1.24} & 1.82 ± 0.24  \\  
    Da. Morph. LLaMa & 10.17 & 24.61 & 49.03 ± 0.62 & -0.84 ± 0.86 & 27.80 ± 0.65 & 0.25 ± 0.02 \\
    Da. Mixed. LLaMa & 12.18 & 29.07 & 49.08 ± 0.58 & -0.14 ± 1.26 & 30.46 ± 1.33 & 0.36 ± 0.20 \\ 
    \midrule
    \end{tabular}
    }
    \caption{Results on BPC, BPT, Linguistic Acceptability (LA) \& Summarization (SUM).}
    \label{tab:llm}
\end{table*}

\subsection{Results on downstream evaluation}

The results of our language modeling experiments are presented in \cref{tab:llm}.

\subsubsection{BPC and BPT}

As shown in \cref{tab:llm}, the results reveal a variety in the BPC and BPT scores. Notably, the \textit{Da. Morph. Cerebras} achieves the best BPC score at 3.01, while the \textit{Std. BPE. Cerebras} model has the best BPT score. Overall, the Cerebras models perform relatively consistently in their evaluations. In contrast, the scores of the LLaMA models stand out. The \textit{Std. Da. LLaMA} model clearly outperforms the other three, achieving a BPC of 4.02 and a BPT of 9.40, setting it apart by a substantial margin. The second-best performing LLaMA model is \textit{Da. BPE. LLaMa}, suggesting that while the model struggles with the new Danish vocabulary, it seems to retain familiarity with the BPE tokenizer architecture. In contrast, \textit{Da. Morph. LLaMa} and \textit{Da. Mixed. LLaMa} appear completely out of balance, indicating difficulty adapting to these new tokenization approaches. It seems that the LLaMa models based on the morphological tokenizers become very weak in its predictions, with an increase in BPC score of \textit{6.15} to \textit{8.16} and an increase in BPT score of \textit{15.21} to \textit{19.67}. We hypothesize that the high scores are due to the fact that only the embedding layers were trained on the LLaMA models. As \cref{tab:llm} shows, the Cerebras models demonstrates a clear advantage when fully fine-tuned with new tokenization methods, adapting well to both the Morph and Mixed tokenization techniques. This may highlight the critical importance of a full fine-tuning process, underscoring that training embeddings alone is insufficient for a model to fully incorporate and leverage new tokenization methods.

\subsubsection{Linguistic Acceptability}

Among the \textit{Cerebras} models, the \textit{Da. Morph Cerebras} model achieved the highest F1 score of 49.61 ($\pm$0.50), outperforming the baseline  model (\textit{Std.~BPE. Cerebras}), which attained an F1 of 33.47 ($\pm$0.33). The \textit{Da.~BPE Cerebras} and \textit{Da.~Mixed Cerebras} models also demonstrated slight improvements with F1 scores of 43.51 ($\pm$2.85) and 45.84 ($\pm$2.39), respectively, suggesting that these tokenization methods contribute positively to the model's ability to discern LA. The achieved improvement in F1 score indicates that the \textit{Da.~Morph Cerebras} model - with its increased Morphological vocabulary compared to the other models - is the best at correctly identifying both acceptable and unacceptable linguistic instances, thereby enhancing the model's overall precision and recall.

In contrast, the \textit{LLaMa} models presented a different pattern. Both \textit{Da.~Morph LLaMa} and \textit{Da.~Mixed LLaMa} achieved F1 scores close to the \textit{Std.~BPE LLaMa} model, with \textit{Da.~Morph LLaMa} at 49.03 ($\pm$0.62) and \textit{Da.~Mixed LLaMa} at 49.08 ($\pm$0.58). The \textit{Da.~BPE LLaMa} model underperformed with an F1 score of 33.28 ($\pm$0.27), indicating limitations in capturing the nuances of Danish grammatical coherence compared to its Cerebras counterpart.

Overall, the morphologically informed models demonstrates good F1 performance within both the \textit{Cerebras} and the \textit{LLaMa} models. However, the generally low MCC results across all models suggest that while F1 scores show some improvement with certain models it still indicates that the models' predictions are close to random guessing as an MCC score of 0 reflects purely random performance. This lack of strong correlation, as seen by MCC values near zero, may be attributed to the relatively small size of the models and the limited training of only the embedding layers in the \textit{LLaMa} models.

\begin{table*}[t]
\centering
\resizebox{\linewidth}{!}{
    \begin{tabular}{l|ccc|ccc|ccc|ccc|c}
    \toprule
    \multicolumn{1}{c|}{} & \multicolumn{3}{c|}{\textbf{Conditional Sentence}} & \multicolumn{3}{c|}{\textbf{Dialog with Emotion}} & \multicolumn{3}{c|}{\textbf{Descriptive Language}} & \multicolumn{3}{c|}{\textbf{Historical Narratives}} & \multicolumn{1}{c}{\textbf{Average}} \\
    \textbf{Model} & F & N & G & F & N & G & F & N & G & F & N & G & Score \\
    \midrule
    Std. BPE. Cerebras & 2,4 & 1,6 & 3,6 & 4,0   & \textbf{4,2} & 4,2 & 2,6 &
    3,4 & 3,4 & \textbf{4,0}   & \textbf{3,2} & 4,0   & 3,4 \\ 
    Da. BPE. Cerebras  & 2,2 & 2,0   & 2,2 & 2   & 3,0            & 3,8 & 3,0    & 3,2 & 3,4 & 2,2          & 1,6          & 2,2 & 2,6 \\ 
    Da. Morph. Cerebras & 3,2 & \textbf{3,0} & 3,8 & 2,4 & 2,2 & 2,8 & 3,6 & 3,0 & 4,0 & 2,8 & 2,2 & 3,4 & 3,0 \\
    Da. Mixed. Cerebras & \textbf{4,6} & 2,0 & \textbf{4,6} & \textbf{5,0} & 2,0 & \textbf{4,6} & \textbf{4,0} & \textbf{4,0} & \textbf{4,6} & \textbf{4,0} & 2,0 & \textbf{4,2} & \textbf{3,8} \\
    \hline
    Da. Morph. LLaMa & 3,4 & \textbf{3,0} & 3,8 & 2,4 & 1,8 & 2,4 & 2,0 & 1,8 & 2,4 & 2,4 & 1,8 & 2,8 & 2,5 \\  
    \bottomrule
    \end{tabular}
}
\caption{Results on Fluency (F), Naturalness (N) and Grammar (G) in the human evaluation experiment.}
\label{tab:HumanEval}
\end{table*}

\subsubsection{Summarization}

The top-performing \textit{Cerebras} model is \textit{Da. BPE Cerebras}, which achieves a BERT score of 33.99 ± 1.24 and a ROUGE-L score of 2.15 ± 0.09. In comparison, the other \textit{Cerebras} models score quite lower, ranging from 10.96 to 17.07, indicating poor summarization performance. A similar trend is observed with the \textit{LLAMA} models. \textit{Da. BPE LLaMa} attains the highest BERT score of 36.54 ± 1.24, while \textit{Std. BPE LLaMa} records the best ROUGE-L score of 2.13 ± 0.33.

Overall, all models demonstrate low performance in both ROUGE-L and BERT metrics. The ROUGE-L scores suggest that none of the models produce summaries closely matching the expected output. Despite the generally low scores in SUM evaluation, the BERT scores reflect a degree of meaningful semantic alignment in the summaries produced by the models. 

Comparing the Morphological Cerebras Tokenizers, the slightly higher BERT score for the \textit{Da. Mixed. Cerebras} indicate that the Mixed tokenizer provides the model with an advantage when adapting to an out of domain task such as summarization. This is most likely due to the BPE fallback mechanism, which provides the Mixed tokenizer with an increased flexibility. 

\subsubsection{Final remarks on downstream evaluation}
Based on the positive results from conducting a full fine-tune of the Cerebras model with the Morphological tokenizer variants, we hypothesize that a similar performance gain would be achieved on a larger model such as the LLaMa 3.2 1B if it had been fully fine-tuned, and we had used a larger dataset. Ultimately, based on the ScandEval benchmark results and BPC/BPT evaluation, Morphological segmentation seem to give the models an advantage, when training all parameters of the model.

\subsection{Human Qualitative Evalution}

The human evaluation experiment in \cref{tab:HumanEval} shows the top-performing model to be \textit{Da. Mixed. Cerebras}, which achieves the highest average score of 3.8 across the Fluency (F), Naturalness (N), and Grammar (G) metrics. 

Specifically, the model consistently achieves the highest scores in Fluency and Grammar across all evaluated categories. However, it struggles with Naturalness in some categories, a limitation that is also observed in most of the other models. This suggests that while the models are effective at producing somewhat smooth and grammatically correct text, they face challenges in generating naturally flowing sentence continuations.

In contrast, the \textit{Da. Morph. LLaMa} model records the lowest average score of 2.5, indicating weaker performance in human evaluations. The remaining Cerebras models show decent performance, with \textit{Std. BPE. Cerebras} and \textit{Da. Morph. Cerebras} attaining average scores of 3.4 and 3.0, respectively. Overall, the results highlight that \textit{Da. Mixed. Cerebras} generates more fluent, natural, and grammatically correct language across various categories, whereas the \textit{Da. Morph. LLaMa} model underperforms in most of these human evaluation metrics.

\section{Analysis}

\begin{table*}
\centering
\resizebox{\linewidth}{!}{
\small
    \begin{tabular}{llllll}
    \toprule
    \textbf{Error Category}  & \textbf{Lang.} & \textbf{Sentence with grammatical error} & \textbf{Correction}   \\
    \midrule
        \multirow{6}{*}{Word order}
        & Da & "Og i kvote II udløber ansøgningsfristen \textbf{15. den} marts 1992." & den 15.  \\
        & Eng & "And in quota II, the application deadline expires on \textbf{of 15th} March 1992." & 15th of  \\
        
        \cline{2-4}

        & Da & "Og rystede, det \textbf{tyskerne var}." & var tyskerne \\
        & Eng & "And shaken, that \textbf{were the Germans}." & the Germans were \\
        
        \cline{2-4}

        & Da & "Alle \textbf{fører veje} til Rom." & veje fører \\
        & Eng & "All \textbf{lead roads} to Rome." & roads lead \\
        
        \hline
        
\multirow{6}{*}{Missing word}
         & Da & "(...) og tog hendes hænder i sine, uden at vide hvad \textbf{\_\_} skulle bruge dem til." & han \\
         & Eng & "(...) and took her hands in his, without knowing what \textbf{\_\_} was supposed to use them for." & he \\

         \cline{2-4}

         & Da & "(...) og at rejse derover som dansk reporter, er \textbf{\_\_} bede om at blive arresteret som spion." & at \\
         & Eng & "(...) and traveling there as a Danish reporter is \textbf{\_\_} ask to be arrested as a spy." &  to \\

         \cline{2-4}

        & Da & "En stor del af Ungbos 1.600 ungdomsboliger \textbf{\_\_} 2.200 hybler er truet af tvangsauktion." & og \\
        & Eng & "A large portion of Ungbo's 1,600 housing units \textbf{\_\_} 2,200 apartments are threatened with foreclosure." & and \\
        \hline
    
    \end{tabular}
    }
    \caption{Typical error categories derived from the LA predictions of Cerebras models and examples of grammatically incorrect sentences and their correction from each category. "(..)" means that a part of the sentence have been left out and "\_\_" has been inserted to emphasize the missing word.}
    \label{tab:laErrors}
\end{table*}

\subsection{Tokenizers}

To identify patterns in tokenizer performance and pinpoint areas where they may struggle with Danish morphological structures, we investigated their ability to segment words in the \textit{Prefix} and \textit{Compounds with linking} categories as these were the ones with the overall lowest scores among the used tokenizers. 

Starting with the Prefix category, the \textit{Da. Morph. LLaMa} tokenizer demonstrated morphological awareness by correctly segmenting prefixes in words like \textit{mistænk} into (mis-tænk; eng: "suspect") and \textit{ugift} into (u-gift; eng: "unmarried"). However, the \textit{Da. Morph. LLaMa} failed to segment the prefixes in words such as \textit{import}, undersegmenting the word by outputting (import; eng: "import"). We hypothesize that this could be due to a general difficulty with Latin-derived prefixes. \textit{The Da. Morph. Cerebras} tokenizer showed similar patterns, accurately segmenting native Danish prefixes but struggling with words of foreign origin. 

The \textit{Da. Mixed} tokenizers showcased mixed results. While they correctly segmented some words with prefixes, they also over-segmented others, such as splitting \textit{bedøm} into (bed-øm; eng: "rate") instead of the correct morphological segmentation (be-døm), and \textit{ekskone} into (ek-sk-one; eng: "ex-wife") instead of the correct (eks-kone), misidentifying the morphological segments. This may suggest that combining BPE with Morfessor can introduce fragmentation when subword units resemble parts of prefixes or roots.

Tokenizers relying solely on BPE, such as the \textit{Da. BPE} variants, struggled with prefix recognition. They often failed to identify the prefix-root boundary, as seen in the segmentation of \textit{benyt} into (ben-yt; eng: "use") instead of (be-nyt), and \textit{bebuder} into (beb-uder; eng: "announces") rather than (be-bud-er). The standard BPE tokenizers performed the poorest, frequently over-segmenting words without regard of morphological structure, indicating a lack of morphological awareness.

For the compounds with linking category, both versions of the morfessor implementations generally performed well, accurately identifying linking elements and segmenting compounds such as \textit{plejehjem} into (plej-e-hjem; eng: "nursing home") and \textit{løbesko} into (løb-e-sko; eng: "running shoes"). They also correctly segmented \textit{landstræner} as (land-s-træn-er; eng: "national coach"), showcasing an understanding of where linking elements should appear within compound words. However, the morphological tokenizers still faced challenges. Words like \textit{verdenskrig} (correct annotation: verden-s-krig; eng: "world war") and \textit{synspunkt} (correct annotation: syn-s-punkt; eng: "point of view") were often segmented without isolating the linking "s". Similarly, while \textit{skrivelyst} (correct annotation: skriv-e-lyst; eng: "urge to write") was sometimes handled correctly, there were instances where the morphological tokenizer were under-segmenting the words, treating them as if they were a root morpheme.

The BPE-only tokenizers struggled more noticeably. They often treated entire compounds as single units or produced arbitrary splits, failing to capture nuanced morphological segments. For instance, the \textit{Da. BPE. LLaMa} and \textit{Da. BPE. Cerebras} tokenizer segmented \textit{plejehjem} as (plejehjem) rather than (plej-e-hjem), and \textit{skrivelyst} as (skri-vel-yst) instead of (skriv-e-lyst). Likewise, \textit{verdenskrig} appeared as (verden-skrig) rather than isolating the linking element “s”. The Standard BPE tokenizers (Std. BPE. LLaMa and Std. BPE. Cerebras) also produced morphologically irrelevant splits, such as (ver-d-ens-k-rig) for \textit{verdenskrig} and (sk-r-ivel-yst) for \textit{skrivelyst}.

An important question to raise, is the previously mentioned mismatch in category representation in the annotated dataset for the Morfessor framework. Although having a larger proportion of root morphemes (150) aligns well with the typical morphological structure of Danish, where root morphemes form the backbone of vocabulary, it also means that categories like Prefixes (38) and Linking Morphemes (40) receive less emphasis during training. This imbalance may lead the Morfessor-based tokenizers to sometimes under-segment and struggle with less frequent or more complex morphological structures.

\subsection{Downstream Evaluation}

The downstream evaluation using Scandeval generally showed promising results for the morphological tokenizers. All models - both LLama and Cerebras - generally present a decent understanding of Danish Linguistic based on LA scores, but perform underwhelming when it comes to specific text generation using the SUM benchmark. For both BPC and BPT scores we did also see a very similar performance across the Cerebras models. Therefore, we find it most relevant to look further into analyzing the results of LA for the fully trained Cerebras models, as these seemed to be improved the most by the Morphological tokenization approach. 

Running the ScandEval LA benchmark with the flags \textit{--verbose} and \textit{--debug} allowed us to access the predictions made by the models, which for the 4 Cerebras models totals to 51.680 predictions. While that is a large amount of predictions, we manually looked for tendencies in the cases of the models predicting false positives - that a sentence is grammatically correct ("ja") when the correct label is "nej". As native Danish speakers, we were in many cases able to identify the reason why the model was wrong in its prediction. 
In this process, we noticed two primary groups of error types, resulting in most wrong predictions from the Cerebras models. These groups are presented in the ScandEval paper, and are defined as \cite{nielsen2023scandeval}:

\begin{itemize}
    \item \textbf{Incorrect word order}: Two words have switched places in the sentence, resulting in a grammatically incorrect sentence.
    \item \textbf{Missing word}: A word has been removed from the sentence, obfuscating the grammatical correctness.
\end{itemize}

The categories are presented in \cref{tab:laErrors} along with 3 examples of sentences with errors and how they could be corrected. When analyzing the type of wrong predictions made by the Cerebras models, it was clear to see that the models made wrong predictions related to the structure of the sentence, such as incorrect word order and missing words.  

Looking at the predictions, we also see examples of the opposite when analyzing the output - that is, the models correctly predict that the sentence is grammatically correct, but ScandEval labels it not to be. Two such examples is: \textit{"Han skulle komme ud til en for at hente noget, inden det blev mørkt"} (eng: "He was going to come out to somebody to fetch something before it got dark") and \textit{"forstår hinanden godt, og jeg nyder at heade ham fri til scoring"} (eng: “understand each other well, and I enjoy setting him up for a free scoring opportunity”). Both sentences are grammatically correct, and it could seem that a word has been removed, without it affecting the grammatical correctness of the sentence. These examples show that the automatic word removal and word order change ScandEval makes use of, in some cases still produces a correct sentence \cite{nielsen2023scandeval}. This will have a negative influence on the results form the evaluation, but how big is difficult to say. 

Conclusively, we also find it important to address the remarkably low LA F1 score for \textit{Da. BPE. LLaMa} compared to the other LLaMa variants. This is unexpected, especially when looking at the BPC, BPT and BERT scores, which indicates that this model is better than the Morph and Mixed LLaMa models. When analyzing the model output during the ScandEval evaluation, it appears that the model predicts "Nej" (eng: No) to most sentences. It has not been possible to identify anything in the ScandEval code that explain this behavior. We hypothesize that this behavior is the result of the combination of the \textit{Da. BPE LLaMa} tokenizer being trained exclusively on Danish text and only adjusting the embeddings of the LLaMa model. This results in the \textit{Da. BPE. LLaMa} model having a strong preference of outputting "Nej" when presented with the sentences from ScandEval. 

\subsection{Human Qualitative Evaluation}

As seen in the evaluation results, the \textit{Da. Mixed. Cerebras} model outperforms all models, including the \textit{Da. Morph. Cerebras} model. This is despite the latter excelling in the LA task in ScanEval as well as achieving the best overall F1 score in identifying different morphological categories. This suggests that a combination of morphological awareness as well as the flexibility of the BPE algorithm enhances its ability to generate fluent and grammatically correct text.

However, an important discussion of this study is the use of out-of-domain data, where the models were exposed to sentence structures and content that they had probably not encountered during training. This missing familiarity may have hindered the models' ability to produce natural continuations, as showcased by the consistently low Naturalness scores across the models. 
As seen when generating model responses for the experiment, they varied depending on how the prompts were worded. Even small changes in the prompts led to different performance results. This means that the outcome we observe might be influenced by prompt-response combinations rather than the models' true abilities.

Moreover, many of the generated sentences included legal text, possibly reflecting the models' training data exposure to such content. The inclusion of domain-specific language indicates that the models may have been exposed to an over-representation of certain data patterns, reducing their performance in generating contextually appropriate continuations for the given tasks. 

Lastly, it is important to consider the number of participants for the evaluation. The small sample size of only five participants in the human evaluation can introduce potential biases and limit the generalization of the findings. This should be taken into account when assessing the results of the experiment.

\section{Conclusion}
\looseness=-1
We introduced a Danish morphological tokenizer, built using an annotated morpheme dataset and a semi-supervised morphological segmentation model, achieving strong performance in morphological segmentation with an F1 score of 58.84, surpassing a commonly used subword tokenization method. Our experiments demonstrate that morphologically informed tokenizers enhance the downstream performance of generative transformer models, particularly in tasks requiring linguistic understanding of Danish text, such as grammatical coherence. To further address our contributions, we turn to the specific sub-questions posed in our study:

\begin{enumerate}
  \setlength\itemsep{0cm}
    
    \item \textit{What effects do semi-supervised and unsupervised learning have on the Morfessor segmentation accuracy in Danish?}
    
    The Morfessor model trained with the largest annotated dataset (400 words) achieved the highest F1 score of 0.73, compared to 0.35 with no annotations, demonstrating a performance gain from annotated data and utilizing a semi-supervised morphological segmentation model.
     
    \item \textit{How does the performance of a morphologically informed tokenizer compare to common subword tokenizers in accurately segmenting Danish words into morphemes?}
 
    The evaluation of the four tokenization methods showed a notable improvement, with the tokenizers based solely on Morphological segmentation (\textit{Da. Morph}) achieving an F1 scores up to 58.84, compared to 39.28 for the Danish-specific BPE tokenizers (\textit{Da. BPE.}).
     
    \item \textit{How does a morphologically informed tokenizer impact generative transformer model performance in downstream tasks for Danish?}
    
    Our findings reveal that tokenizers based on morphological segmentation achieve BPC and BPT scores that are closely aligned with those of the original tokenizer architectures. This demonstrates that adopting a morpheme-based approach does not compromise the compression efficiency of these models.
    Additionally, generative transformer models employing morphological tokenizers outperformed those using Byte-Pair Encoding (BPE) in two out of three downstream tasks: Linguistic Acceptability and Human Evaluation. The \textit{Cerebras-GPT 111M} model based solely on morphological tokenization achieved the highest F1 score for linguistic acceptability at 49.61, compared to 43.51 for the Danish-specific BPE variant. Similarly, the \textit{Cerebras-GPT 111M} model using a combination of morphological and BPE tokenization recorded an average human evaluation score of 3.8, substantially higher than the Danish-specific BPE variant, which scored 2.6.
    
\end{enumerate}

Overall, these findings demonstrate that morpheme-based tokenization offers notable improvement over a common subword tokenization technique when training generative transformer models, effectively addressing the challenges of Danish morphology.

\section*{Acknowledgments}

We extend our deepest gratitude to Anette Jensen for providing us with 821 annotated Danish morphemes. Her contribution made a significant impact on the depth of our study. Lastly, we wish to thank the IT University of Copenhagen (ITU) and LUMI for granting us access to their high-performance computing facilities.

\bibliographystyle{styles/acl_natbib}
\bibliography{main}

\newpage

\end{document}